\documentclass{article} %
\usepackage{iclr2027_conference,times}

\usepackage{amsmath,amsfonts,bm}

\def\eqref#1{equation~\ref{#1}}

\def\1{\bm{1}}

\DeclareMathAlphabet{\mathsfit}{\encodingdefault}{\sfdefault}{m}{sl}
\SetMathAlphabet{\mathsfit}{bold}{\encodingdefault}{\sfdefault}{bx}{n}

\usepackage{graphicx}
\usepackage{wrapfig}
\usepackage{booktabs}
\usepackage{multirow}
\usepackage{array}
\usepackage{xcolor}
\usepackage{listings}
\usepackage{placeins}
\usepackage{hyperref}
\usepackage{url}
\lstdefinestyle{edgeprompt}{
  basicstyle=\ttfamily\scriptsize,
  breaklines=true,
  breakatwhitespace=false,
  columns=fullflexible,
  keepspaces=true,
  showstringspaces=false,
  frame=single,
  rulecolor=\color{black!20},
  backgroundcolor=\color{black!2},
  xleftmargin=0.5em,
  xrightmargin=0.5em,
  aboveskip=0.6em,
  belowskip=0.6em
}
\graphicspath{{figures/}}

\title{EdgeKernelBench: Unified LLM-Driven\\Kernel Generation for Edge Backends}

\author{%
\textbf{Yuhua Zheng\textsuperscript{1,2$*\dagger$}\hspace{0.9em}Xueyi Chen\textsuperscript{1$*$}\hspace{0.9em}Xinze Zou\textsuperscript{2}\hspace{0.9em}Xin Li\textsuperscript{1,2$\dagger$}\hspace{0.9em}Haoji Hu\textsuperscript{2$\ddagger$}\hspace{0.9em}Huan Wang\textsuperscript{1$\ddagger$}} \\
\textsuperscript{1}Westlake University \quad \textsuperscript{2}Zhejiang University \\
\href{mailto:yuhuazheng@zju.edu.cn,haoji_hu@zju.edu.cn}{\texttt{\detokenize{{yuhuazheng,haoji_hu}@zju.edu.cn}}} \quad \href{mailto:wanghuan@westlake.edu.cn}{\texttt{wanghuan@westlake.edu.cn}} \\
\url{https://pillor9.github.io/EdgeKernelBench/}
}

\iclrfinalcopy
\begin{document}

\vspace*{-42pt}
\maketitle
\lhead{Preprint.}
\begingroup
\renewcommand{\thefootnote}{\fnsymbol{footnote}}
\footnotetext[1]{Equal contribution.}
\footnotetext[2]{Work done as a visiting research intern at ENCODE Lab, Westlake University.}
\footnotetext[3]{Corresponding authors.}
\endgroup
\enlargethispage{11pt}
\vspace{-23pt}

\begin{abstract}
Recent works show that LLMs can generate kernels with strong performance on server GPUs. Edge deployment presents distinct challenges because the hardware is heterogeneous, the software stacks are fragmented, and the resource budgets are tight. LLM kernel generation across diverse edge backends has yet to be studied systematically. We introduce \textbf{EdgeKernelBench}, which combines a unified IR deployment framework with a hierarchical benchmark for edge workloads. The unified IR captures reusable computation, scheduling, and memory decisions and connects to each target backend through a compiler interface. The benchmark spans progressively larger program scopes, from individual operators to compact model graphs, across representative edge application domains. We instantiate the design with TVM as a reference compiler and validate it across heterogeneous edge backends. With a single generation attempt, the best model achieves 64.5\% Correct@1 but only 25.5\% $\mathrm{Fast}@1.0$. This gap shows that functional correctness does not consistently translate into performance gains. We therefore introduce the \textbf{Layered Edge Kernel Agent (LEKA)}, which organizes optimization through a layered IR. On the agent evaluation subset, LEKA achieves 43.5\% $\mathrm{Fast}@1.0$, compared with 13.0\% for direct generation and 26.1\% for Claude Code. LEKA and Claude Code use the same base model and optimization budget. Together, EdgeKernelBench and LEKA support unified evaluation and layered optimization of kernels generated by LLMs across edge backends.
\end{abstract}

\section{Introduction}
\label{sec:intro}
Growing demand for low latency, reduced network reliance, and local data processing drives model deployment on edge devices \citep{zhou2019edgeintelligence,xu2021edgeintelligence}. Yet rapidly evolving models must run across diverse runtimes and hardware backends. Deployment systems continue improving inference efficiency across platforms \citep{nvidia_tensorrt,jiang2020mnn,nachin2026executorch}. Still, new operators and backends often require custom kernels. The need to manually implement, validate, and maintain high-performance operators remains a persistent bottleneck in edge deployment.

\begin{figure}[t]
  \centering
  \includegraphics[width=1.0\linewidth]{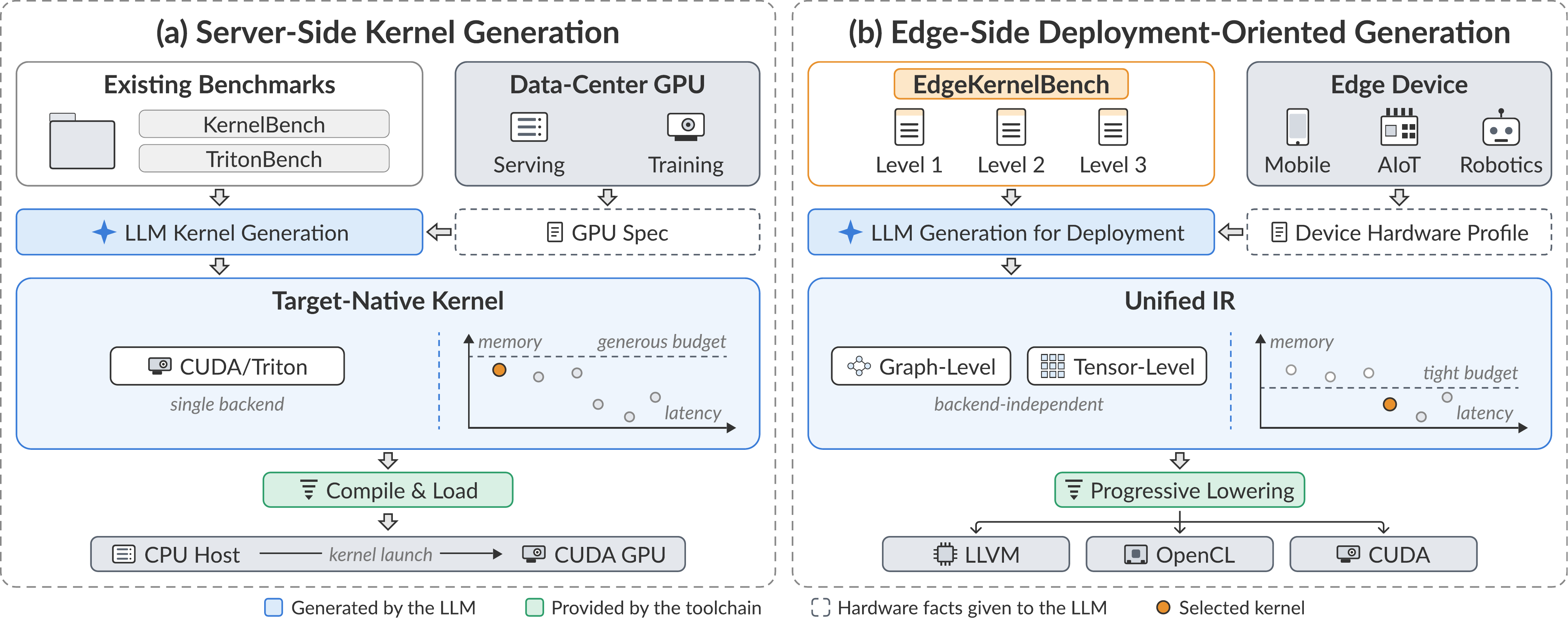}
  \caption{ \textbf{Server and edge kernel generation follow different optimization regimes.} Server methods usually emphasize execution speed within relatively uniform software stacks and generous resource budgets. Edge generation spans fragmented hardware and software ecosystems. It must also balance performance, memory use, and deployment feasibility under tight resource limits.
  }
  \label{fig:edge-generation-scope}
\end{figure}

Large language models can now generate executable code from natural language descriptions \citep{chen2021codex,li2022alphacode}. Recent works apply this capability to high-performance kernel generation \citep{ouyang2025kernelbench,li2025tritonbench,lin2026solexecbench}. However, most of this work still targets server GPUs. There are three major problems that prevent these methods from being used to generate kernels for edge devices, as illustrated in Figure~\ref{fig:edge-generation-scope}.

\noindent\textbf{(1) Edge hardware and software ecosystems are highly fragmented.} Server kernel generation usually targets a small set of mainstream accelerators supported by mature software stacks. Edge devices span heterogeneous processors, vendor runtimes, and deployment engines. Generating native programs directly ties reusable computation and scheduling knowledge to the languages, interfaces, and legality rules of individual backends. Supporting a new backend therefore requires the model to adapt to another programming and deployment environment, which limits transfer across devices.

\noindent\textbf{(2) Edge kernel generation has a different optimization objective from server kernel generation.} Server kernels often run on accelerators with ample dedicated memory, allowing optimization to focus primarily on execution speed. Edge workloads share compute and memory with the rest of the system, and memory capacity can determine whether a model is deployable. Edge kernel generation must therefore optimize performance and memory jointly within a fixed resource budget.

\noindent\textbf{(3) Existing benchmarks do not fully capture the requirements of edge kernel generation.} KernelBench and related studies mainly target server GPUs \citep{ouyang2025kernelbench,li2025tritonbench}, while MobileKernelBench extends evaluation to mobile devices \citep{zou2026mobilekernelbench}. Its public tasks consist mainly of individual operators and their attribute variants. Such tasks assess basic code generation, but not cross-operator optimization or graph-level resource planning in edge deployments.

To address these problems, we introduce \textbf{EdgeKernelBench}. It combines a unified deployment framework with a hierarchical benchmark for edge workloads. The LLM does not produce a separate native program for each backend. The LLM expresses optimization choices for the target device in a unified intermediate representation, or IR. A deterministic backend adapter then lowers the program to the target backend in stages. This separation isolates reusable computation and knowledge from interfaces that are specific to each backend. EdgeKernelBench also places scheduling and memory decisions in the same optimization space. The model can reuse optimization knowledge across backends while balancing execution performance and memory cost under deployment constraints.

All EdgeKernelBench tasks are PyTorch reference programs with traceable provenance \citep{paszke2019pytorch}. The tasks are organized along two orthogonal axes, compositional complexity and application domain. The complexity axis contains three levels named \emph{Basic Compute Patterns}, \emph{Composed Operator Graphs}, and \emph{Compact Model Graphs}. The application domains cover Mobile and Personal Intelligence, Embodied Robotics, and Smart AIoT. Separating these axes enables targeted evaluation of optimization capabilities across edge deployment settings.

To test the feasibility of this design, we instantiate EdgeKernelBench with TVM as the reference compiler stack \citep{chen2018tvm}. We evaluate five LLMs on 650 task instances across CPU/LLVM, GPU/CUDA, and GPU/OpenCL execution paths. The results show that a consistent workflow for generation, validation, and measurement runs across heterogeneous backends. The hierarchical benchmark also reveals how the model capability changes as the program scope grows.

The experiments also expose the limits of one-shot generation. Across 650 task instances, the strongest model achieves 73.5\% Compile@1 and 64.5\% Correct@1, but only 25.5\% $\mathrm{Fast}@1.0$. At Level~3, the highest $\mathrm{Fast}@1.0$ scores on the three backends fall further to values between 0.0\% and 4.0\%. As the program scope grows, one-shot generation struggles to coordinate increasingly coupled decisions across representation levels.

To address this limitation, we further introduce \textbf{LEKA}, the \textbf{L}ayered \textbf{E}dge \textbf{K}ernel \textbf{A}gent. LEKA decomposes coupled cross-level decisions along the IR hierarchy, allowing each agent to focus on one representation while preserving coordination across levels. On a randomly sampled 69-instance subset for agent evaluation, LEKA raises Correct@1 from 42.0\% to 100.0\% and $\mathrm{Fast}@1.0$ from 13.0\% to 43.5\% relative to one-shot generation. With the same base model, task contract, and wall-clock budget, LEKA raises $\mathrm{Fast}@1.0$ from 26.1\% for the pure Claude Code baseline to 43.5\%.

Our work makes three contributions.
\begin{itemize}
    \item We introduce EdgeKernelBench, which combines a unified IR deployment framework with a hierarchical benchmark for edge workloads. It provides a common basis for studying LLM kernel generation and optimization on edge devices.
    \item We introduce LEKA, a general agent architecture for edge kernel generation and optimization. LEKA organizes optimization through a layered IR to reduce the complexity of decisions across representation levels.
    \item We conduct systematic experiments on a range of representative edge hardware backends. The evaluation measures the edge kernel generation capabilities of leading LLMs and examines their main failure modes. It also tests the optimization gains from LEKA.
\end{itemize}

\section{Related Work}
\label{sec:related}

\paragraph{Kernel generation benchmarks.}
Benchmarks have spread from server GPUs to NPUs, TPUs, and multi-vendor accelerators \citep{wen2026multikernelbench,npukernelbench2026,kernelgenbench2026}, but the hardware stays largely server-class, the tasks stay server-oriented, and the artifact stays native code \citep{yu2026kernelgensurvey}. One benchmark reaches mobile devices, and it covers mainly single operators and their attribute variants \citep{zou2026mobilekernelbench}. Native code leaves the model to meet each backend's language and legality rules itself, so correctness collapses on hardware it has rarely seen \citep{wen2026multikernelbench,kernelgenbench2026}. These benchmarks mostly share the KernelBench protocol, which organizes \mbox{PyTorch} reference programs by scope and scores generated kernels on correctness and speedup over a software baseline \citep{ouyang2025kernelbench}. EdgeKernelBench keeps the protocol but evaluates on heterogeneous edge backends, draws its tasks from edge applications, spans single operators to compact model graphs, and replaces native code with one IR shared by all backends.

\paragraph{Compilers and portable kernel representations.}
Deployment stacks fix their optimization knowledge in expert-written kernels or schedule rules, whereas LLM generators hold such knowledge but emit native code and forgo the portability a compiler provides. Inference engines ship a hand-written kernel for each operator and backend \citep{jiang2020mnn,david2021tensorflow,nvidia_tensorrt}. Tensor compilers lower one program through several IR levels to many backends \citep{chen2018tvm,lai2025relax,lattner2021mlir} and search a rule-defined schedule space with cost models and hundreds to thousands of on-device measurements \citep{zheng2020ansor,shao2022metaschedule}. Neither can reach a transformation outside its encoding, such as an algebraic rewrite or an unforeseen fusion boundary. EdgeKernelBench puts the LLM's computation, scheduling, and memory decisions in a unified IR that a deterministic backend adapter lowers to each backend.

\paragraph{LLM-driven kernel optimization.}
Methods beyond one-shot generation differ in where they spend compute but most treat one native program as the unit of optimization. Training-time methods fine-tune a model with execution rewards until it specializes in one kernel language \citep{baronio2026kevin,bai2026dice,li2025autotriton,npukernelbench2026}. Inference-time methods keep the model fixed and iterate through agent roles with profiler feedback \citep{wei2025astra,zhang2025cudaforge}, evolutionary search over a candidate lineage \citep{chen2026avo}, repair loops inside a mobile framework \citep{zou2026mobilekernelbench}, or a generate-test-refine loop over a kernel abstraction layer \citep{saba2026cutegen}. That unit entangles algorithm, fusion, thread-mapping, and instruction choices, so a reward scores the whole program and a profile points at a slow kernel, not at the decision level behind it. LEKA gives each IR level its own agent and routes every measurement to the level that owns the decision rather than to the whole program.

\section{EdgeKernelBench}
\label{sec:edgekernelbench}

\begin{figure}[!t]
  \centering
  \includegraphics[width=\linewidth]{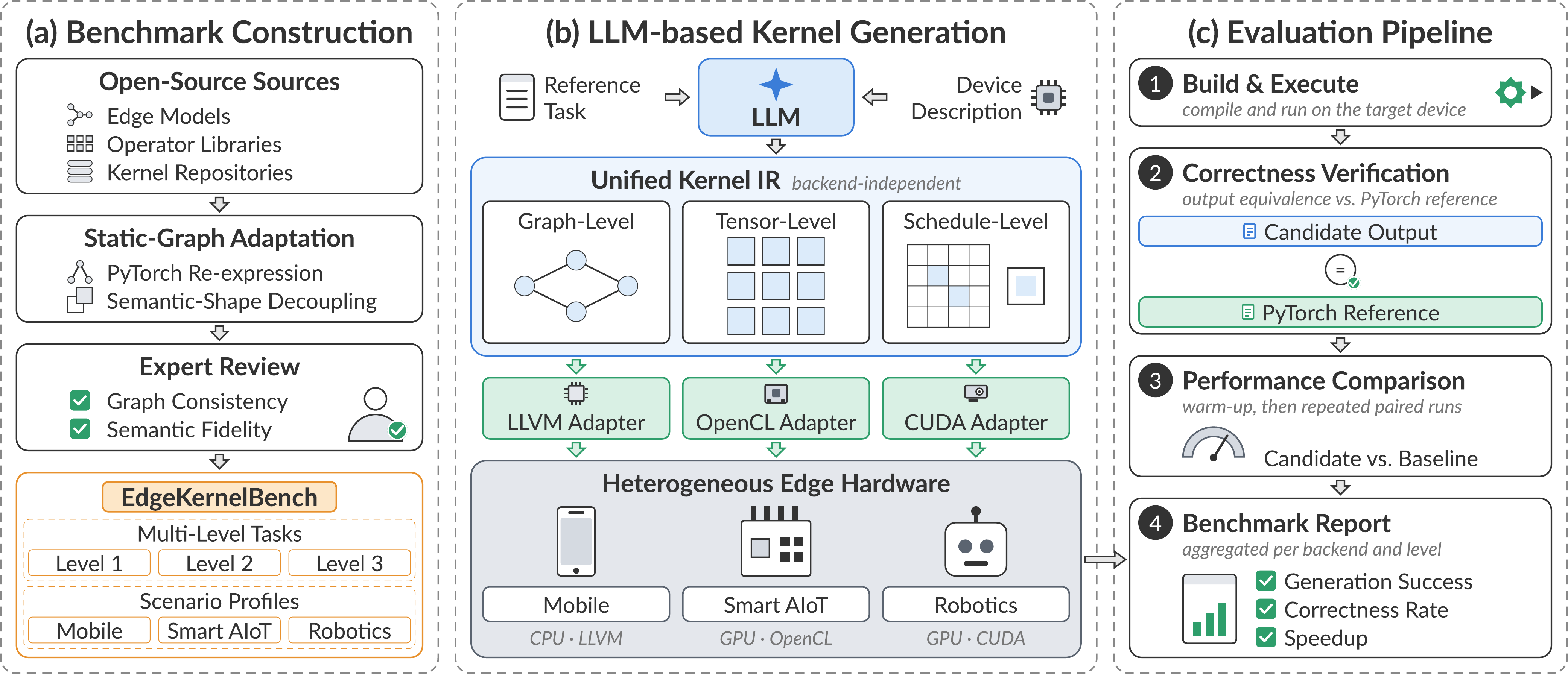}
  \caption{\textbf{Overview of EdgeKernelBench.} \textbf{(a)} We curate workloads relevant to edge deployment as PyTorch tasks and organize them by compositional complexity and application domain. \textbf{(b)} The LLM uses a layered generation interface to express decisions about graphs, scheduling, and memory. A concrete compiler implementation maps these decisions to execution paths. \textbf{(c)} Each candidate is built, checked against a PyTorch reference program, and timed against a paired baseline.}
  \label{fig:edgekernelbench-overview}
\end{figure}

\subsection{Benchmark Construction and Organization}
\label{sec:benchmark-construction}

\noindent\textbf{Benchmark construction.} Figure~\ref{fig:edgekernelbench-overview}(a) shows the three-stage construction of EdgeKernelBench. We first select basic computations, model modules, and operator compositions representative of practical edge inference. We draw these tasks from widely used open-source edge models, operator libraries, and kernel repositories. Then, we convert dynamic computation into static form and give every task a common PyTorch interface. This step also separates the core computational semantics from the workload configuration used by a specific application. Finally, we manually inspect each graph for structural consistency and intended semantics.

\noindent\textbf{Organization along two dimensions.} EdgeKernelBench contains 331 canonical tasks organized along two orthogonal dimensions. The first dimension captures compositional complexity. \emph{Level~1 Basic Compute Patterns} contains 128 basic operators or core computation patterns. It tests whether an LLM can preserve local semantics, produce a legal implementation, and make basic scheduling decisions for the target. \emph{Level~2 Composed Operator Graphs} contains 162 fused computation graphs or model modules. It tests whether the model can coordinate dependent operators, intermediate results, and fusion choices beyond a single operator. \emph{Level~3 Compact Model Graphs} contains 41 complex modules or compact model graphs. It evaluates execution planning and correctness over longer dependency chains, branches, and larger program scopes. The second dimension groups tasks by edge application. It includes Mobile and Personal Intelligence with 204 instances, Smart AIoT with 228, and Embodied Robotics with 218. Each domain spans all three complexity levels. Every task corresponds to a meaningful model structure from an edge application. Correctness and performance gains on these tasks therefore indicate potential value for practical edge deployment.

\begin{figure}[!t]
  \centering
  \includegraphics[width=\linewidth]{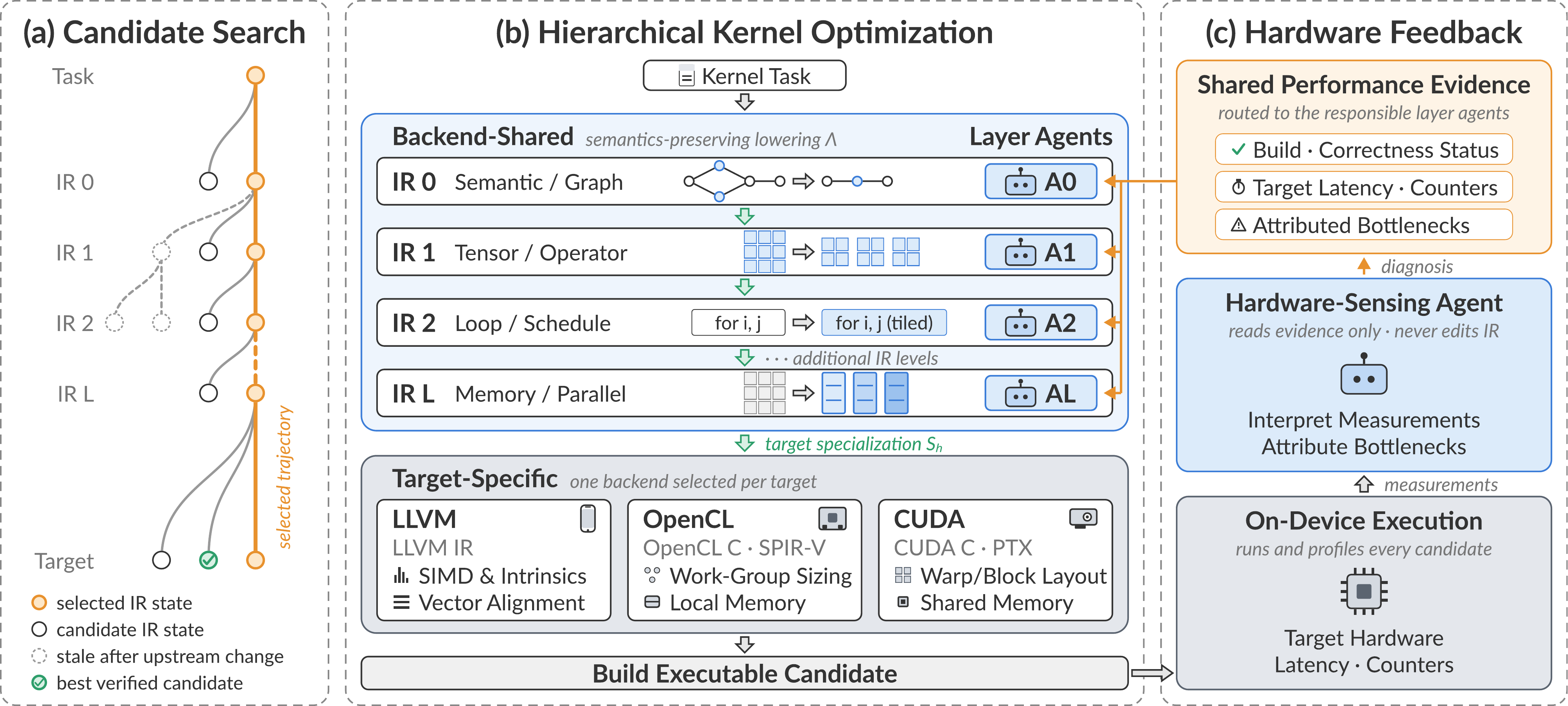}
  \caption{\textbf{Overview of LEKA.} LEKA organizes layer agents along progressive IR lowering. Their scope narrows from the global program structure to implementation details for a specific target. Compiler conversions preserve prior optimizations while opening search spaces at lower levels. The candidate tree stores IR candidates at different levels and records how they were derived. Hardware feedback guides subsequent search at the relevant level.}
  \label{fig:hierarchical-agent}
\end{figure}

\subsection{Unified Generation and Evaluation Framework}
\label{sec:edge-pipeline}

\noindent\textbf{Shared representation.} Edge deployment spans fragmented software stacks and must meet strict resource limits. Program generation must therefore accommodate different backends while balancing execution performance against memory use. EdgeKernelBench separates the shared task and evaluation contract from the deployment implementation for each backend. Given a PyTorch reference program and a description of the target device, the LLM expresses its optimization decisions in \emph{Unified Kernel IR}. The IR places computation, scheduling, and memory in a shared optimization space while preserving target-specific choices. Backend adapters lower the shared IR into target-specific executable programs. This separation makes optimization intent reusable across edge devices and software stacks. New backends can join the same generation and evaluation workflow through dedicated adapters without changing the shared task contract or evaluation protocol.

\noindent\textbf{Unified evaluation pipeline.} EdgeKernelBench provides a unified evaluation of LLM-generated edge kernels across heterogeneous backends. The protocol proceeds through candidate construction, functional validation, and performance measurement. These stages reveal deployability and locate failures throughout the process. The evaluation characterizes each candidate by its compilation success, functional correctness, and execution performance. Its unified structure makes the protocol broadly applicable across diverse edge hardware, compiler stacks, and runtime environments.

\section{Layered Edge Kernel Agent (LEKA)}
\label{sec:agent}

EdgeKernelBench provides a unified generation and evaluation interface, yet one LLM call must coordinate decisions across several abstraction levels. For complex tasks, a single agent must retain a global view of the computation graph while reasoning about implementation details at lower levels within a limited context. As program scope grows, this context burden constrains optimization and makes execution feedback harder to attribute. We therefore introduce the \textbf{L}ayered \textbf{E}dge \textbf{K}ernel \textbf{A}gent, or \textbf{LEKA}. As shown in Figure~\ref{fig:hierarchical-agent}, LEKA decomposes optimization by IR level. A shared candidate tree links decisions across levels and gives each layer agent a focused view of the optimization space exposed at its IR level, while hardware feedback drives subsequent iterations.

\subsection{Hierarchical IR Optimization}
\label{sec:hierarchical-ir-agent}

\noindent\textbf{Agents aligned with representation levels.} Let $q$ denote a kernel task, $h$ the target device, and $\mathcal{X}_{\ell}$ the set of legal program states at IR level $\ell$. Agent $A_{\ell}$ optimizes the representation at level $\ell$ within the semantic boundary and edit scope defined by contract $C_{\ell}$.

At candidate version $t$, LEKA maintains a candidate tree $\mathcal{T}_t=(V_t,F_t)$, where $V_t$ contains all program states generated so far, $F_t$ contains the transitions produced by within-level optimization or cross-level lowering, and $v_{\ell,t}\in V_t$ denotes the current candidate node at IR level $\ell$. Projection $\Pi_{\ell,h}$ extracts the search history and measurement evidence relevant to the current level and target device. The hardware-sensing agent profiles the candidate on the target device and returns the detected bottlenecks to the current level in a report $r_t$. We write optimization within a level and conversion between levels as
\begin{equation}
  \begin{aligned}
    x_{\ell,t+1}
    &= A_{\ell}\!\left(
       x_{\ell,t}\mid q,h,C_{\ell},
       \Pi_{\ell,h}(\mathcal{T}_t,v_{\ell,t}),r_t
       \right), \\
    x_{\ell+1,t}
    &= \Lambda_{\ell\rightarrow\ell+1}\!\left(x_{\ell,t}\right).
  \end{aligned}
  \label{eq:layer-agent-update}
\end{equation}
The compiler supplies $\Lambda_{\ell\rightarrow\ell+1}$, which preserves semantics, performs required legalization and interface conversion, and carries existing IR optimizations forward. It does not search for candidates, leaving performance decisions to the layer agents.

\noindent\textbf{Progressive lowering and focused context.} LEKA follows progressive lowering across multiple IR levels~\citep{lattner2021mlir}. As lowering proceeds, decisions made at upper levels gradually become fixed while finer optimization spaces open. This order lets each decision be optimized at the IR level that expresses it most directly. The scope of each agent narrows in parallel from the full program to local implementation details. Compared with a single agent that must retain both global program structure and low-level details in the same context, LEKA decomposes a complex task into optimization steps aligned with individual IR levels. This decomposition reduces the context required by each call and makes execution feedback easier to attribute to the relevant level.

\noindent\textbf{Shared abstractions and target specialization.} LEKA first searches EdgeKernelBench's backend-shared IR layers, each covering optimization at a different program granularity. During target-specific lowering, the adapter retains high-level optimizations and exposes more hardware-coupled choices to lower-level agents. This separation allows LEKA to transfer high-level optimizations to new backends and focus the remaining adaptation search on target-specific implementation details.
\subsection{Candidate Search Guided by Evidence}
\label{sec:evidence-guided-search}

\noindent\textbf{Candidate search with lineage tracking.} LEKA organizes its feedback loop as a lineage search over the candidate tree introduced above~\citep{yin2025godelagent}. With every optimization attempt, LEKA adds a new state while retaining earlier candidates and their validation results. The orange path in Figure~\ref{fig:hierarchical-agent} marks the active optimization trajectory, while the remaining branches preserve alternatives for comparison, rollback, or continued search. Each layer agent expands candidates within its assigned optimization space, and the shared tree connects these local searches across IR levels.

\noindent\textbf{Feedback from hardware evidence.} For each executable candidate $v_t$, the hardware-sensing agent runs the evaluator on the target device and collects detailed feedback. It uses these measurements to identify the level $\ell_t$ that needs further optimization and writes a report $r_t$ for the candidate. The selected layer agent uses the report attached to the node to generate the next candidate, forming a layered iteration driven by device measurements.

\section{Experiments}
\label{sec:experiments}

\subsection{Experimental Setup}
\label{sec:experimental-setup}

\noindent\textbf{Evaluation environment.} We instantiate EdgeKernelBench and LEKA with TVM 0.23.0\footnote{\url{https://tvm.apache.org/}} \citep{chen2018tvm}, chosen for its mature open-source ecosystem and established use in practice. Through a modified TVM integration, EdgeKernelBench manages task construction, device descriptions, and the optimization workflow. LEKA optimizes computation graphs in Relax and computation and scheduling in TIR. It refines implementations in CUDA C~\citep{nvidia2024cuda126}, OpenCL C~\citep{opencl}, or LLVM IR~\citep{lattner2004llvm}. All optimization decisions come from the LLM or agent. TVM handles representation conversion, target code generation, and runtime support.

We evaluate three devices to cover different edge hardware and software ecosystems. CPU/LLVM runs on an iQOO Neo9S Pro+ with a Snapdragon 8 Gen 3 SoC. The GPU/CUDA device is an NVIDIA Jetson AGX Orin 64GB. For GPU/OpenCL, we use a Firefly ROC-RK3588S-PC with a Mali-G610 GPU. We evaluate compilation, correctness, and performance on each device.

\noindent\textbf{Metrics.} We report compilation success rate (CSR, Compile@1), functional correctness rate (FCR, Correct@1), $\mathrm{Fast}@p$, and geometric mean speedup (GM). CSR is the fraction of tasks whose generated programs compile successfully. FCR additionally requires all output checks to pass. $\mathrm{Fast}@p$ is the fraction of tasks with correct programs, valid timings, and a median paired speedup above $p$ over the baseline \citep{ouyang2025kernelbench}. GM is the geometric mean of speedups over the specified baseline for valid programs on the same task set, including slower results.

\noindent\textbf{Evaluation tasks and models.} Following the EdgeKernelBench configuration, CPU/LLVM uses tasks from Mobile and Personal Intelligence, GPU/CUDA from Embodied Robotics, and GPU/OpenCL from Smart AIoT. We adapt input shapes to each scenario and retain all three complexity levels. Our one-shot evaluation covers representative proprietary models, including GPT-5.6-sol \citep{openai2026gpt56}, Gemini-3.7-Flash \citep{google2026gemini37}, and Claude Sonnet 5 \citep{anthropic2026sonnet5}. We also evaluate the open-weight models DeepSeek-V4-Pro-0813 (abbreviated as DeepSeek-V4-Pro) \citep{deepseek2026v4} and GLM-5.3 \citep{zai2026glm53}. Given time constraints, we use stratified random sampling by complexity level to select 69 task instances across the three backends for LEKA evaluation. LEKA and the pure Claude Code baseline both use GLM-5.3-Flash \citep{zai2026glm53flash} as the base model. They share the same task contracts, evaluator, and time budgets.

\begin{table}[!t]
\centering
\setlength{\abovecaptionskip}{0pt}
\caption{\textbf{One-shot generation results against the TVM baseline.} CSR, FCR, and F@1 denote Compile@1, Correct@1, and $\mathrm{Fast}@1.0$. Percentages cover all tasks in each backend--level slice; $N_{\mathrm{L1/L2/L3}}$ gives task counts. Bold and underlining mark the best and second-best results.}
\label{tab:tvm-baseline-results}
\footnotesize
\setlength{\tabcolsep}{2.4pt}
\renewcommand{\arraystretch}{0.84}
\setlength{\aboverulesep}{0.25ex}
\setlength{\belowrulesep}{0.40ex}

\begin{tabular*}{\linewidth}{@{\extracolsep{\fill}}l*{9}{c}@{}}
\toprule
\multirow{2}{*}{\textbf{Model}}
& \multicolumn{3}{c}{\textbf{Level 1}}
& \multicolumn{3}{c}{\textbf{Level 2}}
& \multicolumn{3}{c}{\textbf{Level 3}} \\
\cmidrule(lr){2-4}\cmidrule(lr){5-7}\cmidrule(lr){8-10}
& CSR & FCR & F@1 & CSR & FCR & F@1 & CSR & FCR & F@1 \\
\midrule
\multicolumn{10}{c}{\textit{CPU/LLVM}\quad $N_{\mathrm{L1/L2/L3}}=91/93/20$} \\
\cmidrule(lr){1-10}
GPT-5.6-sol      & \underline{87.9} & 82.4 & \textbf{46.2} & 68.8 & \underline{62.4} & \textbf{15.1} & \underline{60.0} & \underline{40.0} & 0.0 \\
Gemini-3.7-Flash & \underline{87.9} & \underline{84.6} & \underline{37.4} & \underline{72.0} & 58.1 & \underline{11.8} & \textbf{75.0} & \textbf{45.0} & 0.0 \\
Claude Sonnet 5  & 74.7 & 67.0 & 29.7 & 46.2 & 29.0 & 2.2 & 45.0 & \underline{40.0} & 0.0 \\
DeepSeek-V4-Pro  & \textbf{97.8} & \textbf{94.5} & \underline{37.4} & \textbf{83.9} & \textbf{75.3} & 3.2 & 55.0 & 35.0 & 0.0 \\
GLM-5.3          & 61.5 & 42.9 & 12.1 & 50.5 & 37.6 & 5.4 & 50.0 & \underline{40.0} & 0.0 \\
\midrule
\multicolumn{10}{c}{\textit{GPU/CUDA}\quad $N_{\mathrm{L1/L2/L3}}=98/95/25$} \\
\cmidrule(lr){1-10}
GPT-5.6-sol      & \textbf{81.6} & \textbf{79.6} & \textbf{51.0} & \textbf{73.7} & \textbf{65.3} & \textbf{13.7} & \underline{40.0} & \underline{16.0} & \textbf{4.0} \\
Gemini-3.7-Flash & 68.4 & 66.3 & \underline{44.9} & 43.2 & \underline{40.0} & \underline{11.6} & \textbf{68.0} & \textbf{32.0} & 0.0 \\
Claude Sonnet 5  & 52.0 & 49.0 & 32.7 & 12.6 & 9.5 & 0.0 & 12.0 & 8.0 & 0.0 \\
DeepSeek-V4-Pro  & \underline{77.6} & \underline{69.4} & 34.7 & \underline{57.9} & 35.8 & 4.2 & 16.0 & 12.0 & 0.0 \\
GLM-5.3          & 59.2 & 39.8 & 21.4 & 44.2 & 24.2 & 4.2 & 16.0 & 4.0 & 0.0 \\
\midrule
\multicolumn{10}{c}{\textit{GPU/OpenCL}\quad $N_{\mathrm{L1/L2/L3}}=98/105/25$} \\
\cmidrule(lr){1-10}
GPT-5.6-sol      & \textbf{82.7} & \textbf{73.5} & \textbf{38.8} & \textbf{70.5} & \textbf{55.2} & 7.6 & \underline{28.0} & \underline{16.0} & 0.0 \\
Gemini-3.7-Flash & 67.3 & 60.2 & 32.7 & 57.1 & \underline{49.5} & \textbf{16.2} & \textbf{56.0} & \textbf{32.0} & 0.0 \\
Claude Sonnet 5  & 53.1 & 46.9 & 28.6 & 12.4 & 10.5 & 0.0 & 4.0 & 4.0 & 0.0 \\
DeepSeek-V4-Pro  & \underline{81.6} & \underline{65.3} & \underline{33.7} & \underline{61.0} & 41.0 & 5.7 & 20.0 & 4.0 & 0.0 \\
GLM-5.3          & 64.3 & 35.7 & 15.3 & 47.6 & 33.3 & \underline{11.4} & 20.0 & 8.0 & \textbf{4.0} \\
\bottomrule
\end{tabular*}
\par\vspace{-3pt}
\end{table}

\subsection{Baseline Evaluation}
\label{sec:baseline-evaluation}

\subsubsection{Compiler-Baseline Results}

We use the TVM compilation pipeline tuned with MetaSchedule as the baseline. We evaluate compilation success, functional correctness, and execution performance under one-shot generation. Table~\ref{tab:tvm-baseline-results} shows how generation quality varies with task complexity, hardware backend, and model.

Across task levels, CSR, FCR, and F@1 generally decline as tasks expand from single operators to operator compositions and multi-module computation graphs. Across the three backends, the highest F@1 falls from 51.0\% at Level~1 to 16.2\% at Level~2 and 4.0\% at Level~3. More complex tasks introduce longer semantic dependency chains. They also require models to coordinate computation and storage across a larger program context. These demands make it harder to maintain program consistency and optimize across operators. As a result, generation capabilities on basic operators do not readily extend to complex computation graphs.

Hardware backends also affect generation quality, with generally higher FCR on CPU than on GPU. At Level~1, the highest FCR reaches 94.5\% on CPU, compared with 79.6\% on CUDA and 73.5\% on OpenCL. Thread partitioning and parallel mapping make valid GPU implementations harder to generate. Across the two GPU configurations, CUDA achieves higher F@1 overall. The advantage is most consistent at Level~1, where every model outperforms its OpenCL result. We hypothesize that this partly reflects greater exposure to high-quality CUDA implementations during training.

Models also differ in their strengths. Among proprietary models, GPT maintains relatively high F@1 across multiple backends. Gemini performs well on complex tasks, achieving the highest Level~3 FCR on all three backends. Among open-weight models, DeepSeek achieves strong correctness on CPU, reaching 75.3\% FCR at Level~2. Its corresponding F@1 is only 3.2\%, compared with 15.1\% for GPT. Open-weight models can therefore exceed proprietary models in correctness under some configurations, while still trailing in execution performance. Model strengths in program correctness do not necessarily carry over to performance optimization.

\subsubsection{Production-Inference-Engine Results}
\label{sec:mnn-evaluation}

We compare one-shot LLM-generated programs with two production inference engines, MNN 3.3.0\footnote{\url{https://github.com/alibaba/MNN}} \citep{jiang2020mnn} and TensorRT 10.3.0\footnote{\url{https://developer.nvidia.com/tensorrt}}. Given operator and backend limitations, each comparison uses a fixed task subset that the engine supports and executes correctly. We measure generated programs and their corresponding engine implementations on the same target device.

\edef\edgekernelsavedintextsep{\the\intextsep}
\begin{figure}[!t]
\centering
\includegraphics[width=\linewidth]{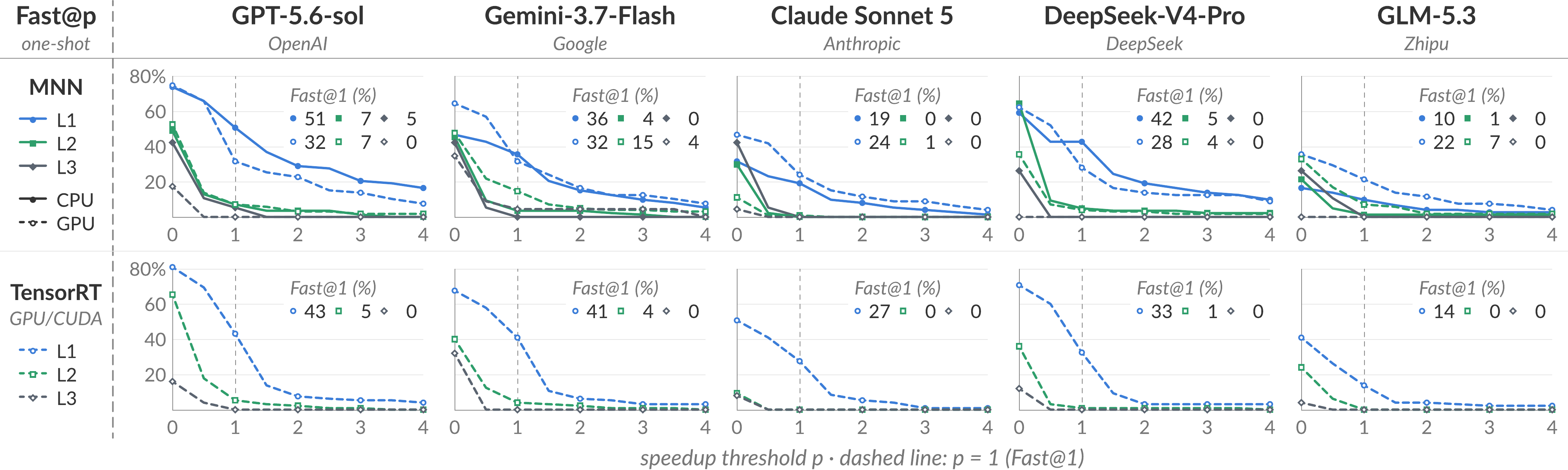}
\caption{\textbf{Performance of one-shot LLM-generated programs relative to MNN and TensorRT.} The horizontal axis shows the speedup threshold $p$, and the vertical axis reports $\mathrm{Fast}@p$ on the fixed task subsets. At $p=0$, the score is the fraction of tasks with correct outputs and valid timings.}
\label{fig:mnn-fast-profiles}
\end{figure}

Figure~\ref{fig:mnn-fast-profiles} shows that programs outperforming MNN are concentrated at Level~1. F@1 peaks at 50.7\% on CPU and 31.6\% on GPU. The maxima at Levels~2 and~3 fall to 14.9\% and 5.3\%, respectively. As the speedup threshold increases, curves for complex tasks rapidly approach zero, with large speedups confined to a few tasks. Speedups also differ between CPU and GPU. These differences may reflect how effectively generated programs use CPU vectorization and GPU parallelism.

Results against TensorRT show a similar pattern across task levels. The highest F@1 is 43.2\% at Level~1 and falls to 5.3\% at Level~2. No Level~3 candidate outperforms TensorRT. The Level~1 curves also decline sharply once the speedup threshold exceeds~1. At a threshold of~2, the highest task fraction is only 7.4\%. One-shot generation can therefore outperform TensorRT on some basic operators, but large speedups remain limited to a small fraction of tasks.

\subsubsection{Error Analysis}
\label{sec:error-analysis}
\setlength{\intextsep}{6pt plus 2pt minus 2pt}
\begin{wrapfigure}{r}{0.34\linewidth}
\centering
\includegraphics[width=\linewidth]{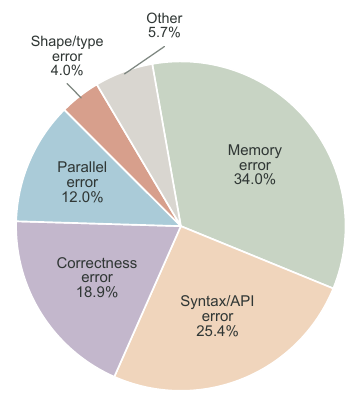}
\caption{\textbf{Failure distribution.} Percentages pool failures across all models and backends.}
\label{fig:error-distribution}
\end{wrapfigure}
We group failures into six categories. \textbf{Syntax/API errors} cover invalid syntax and incorrect interface use. \textbf{Shape/type errors} involve incompatible tensor dimensions or data types. \textbf{Memory errors} concern invalid buffer allocation or access. \textbf{Parallel errors} violate thread mapping or synchronization constraints. \textbf{Correctness errors} indicate output mismatches or consistency violations. \textbf{Other} includes timeouts and unresolved compiler, execution, or verification exceptions.

\smallskip Figure~\ref{fig:error-distribution} pools failures across all models and backends. Memory and syntax/API errors account for 34.0\% and 25.4\% of failures, respectively. Most memory errors concern allocation legality, not insufficient capacity. This pattern suggests limited understanding of buffer scope and storage rules in the target representation. Correctness and parallel errors contribute another 18.9\% and 12.0\%. These failures expose weaknesses in preserving computational semantics and reasoning about dependencies across parallel threads. Overall, these failures reveal limitations in program construction and semantic preservation, informing future edge kernel generation systems. \par \WFclear

\begin{figure}[t]
\centering
\includegraphics[width=\linewidth]{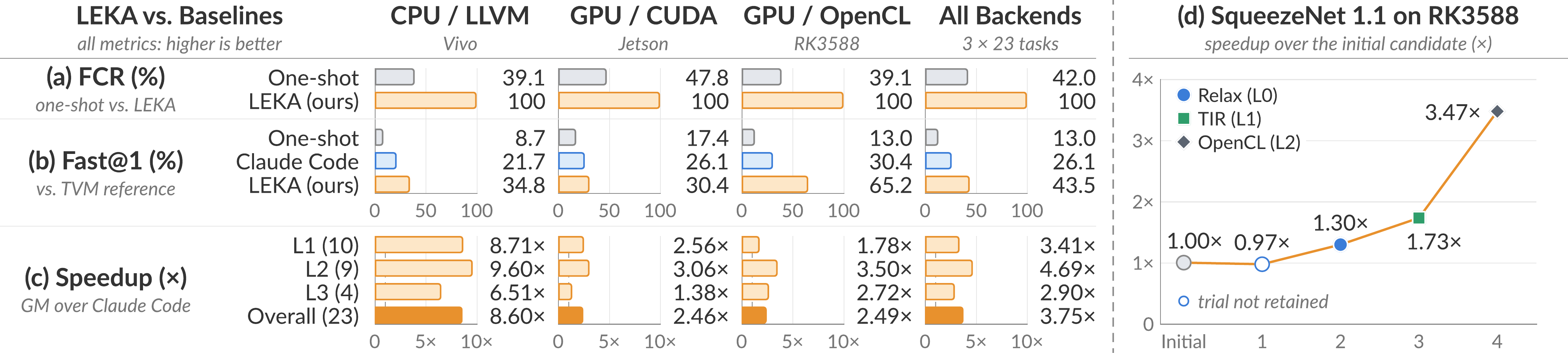}
\caption{\textbf{LEKA results and layered optimization.} (a) FCR and (b) F@1 over TVM on 23 tasks per backend. (c) GM relative to pure Claude Code. (d) Four candidates for SqueezeNet1.1, with candidate-evaluation speedups relative to the initial implementation.
}

\label{fig:leka-results}
\end{figure}

\subsection{LEKA-Guided Optimization}
\label{sec:agent-evaluation}

\subsubsection{Results against One-Shot Generation and Claude Code}
Figure~\ref{fig:leka-results}(a,b,c) compares LEKA with one-shot generation and pure Claude Code across 69 task instances. Relative to one-shot GLM-5.3-Flash, LEKA raises CSR from 60.9\% to 100.0\% and FCR from 42.0\% to 100.0\%. F@1 over TVM increases from 13.0\% to 43.5\%. These final-program scores show the benefit of iterative generation guided by device feedback.

The agent comparison uses the same base model, task contracts, evaluator, and assigned task-level time budgets. LEKA raises F@1 from 26.1\% for Claude Code to 43.5\%, with an overall GM of $3.8\times$ relative to Claude Code. Across all three backends, GM is higher at Level~2 than at Level~1, with pooled GM rising from $3.4\times$ to $4.7\times$. This pattern may reflect LEKA's layered design and the focused scope of each agent. As more operators are composed, optimization requires longer contexts and involves more interdependent decisions. LEKA distributes these decisions across representation levels, allowing each agent to focus on optimization within its own level. This reduces the burden of handling cross-level details and may explain the larger relative gains at Level~2. At Level~3, GM declines to $2.9\times$, possibly reflecting the base model's limited reasoning and optimization capabilities on larger graphs with longer dependency chains. Overall, layered optimization helps manage task complexity, but its gains remain constrained by the base model's capabilities.

\subsubsection{Optimization Case Study}
The SqueezeNet1.1 case on RK3588 illustrates how LEKA's agents coordinate cross-level optimization (Figure~\ref{fig:leka-results}(d)). The Relax agent adopts NHWC layout and rewrites $1\times1$ convolutions as matrix multiplications, exposing opportunities for lower-level optimization. Building on this graph, the TIR agent uses shared-memory tiling to reuse convolution inputs, increasing the recorded speedup from $1.3\times$ to $1.7\times$. The OpenCL agent applies register blocking to the matrix multiplications introduced in Relax while preserving the TIR-optimized convolutions. This refinement reduces repeated input loads and accumulates in registers, bringing speedup over the initial implementation to $3.5\times$. These cumulative gains illustrate how LEKA combines focused reasoning with coordinated optimization across layers. Layered decomposition reduces the reasoning burden of complex optimization, while agent cooperation lets complementary improvements build on one another. \par
\setlength{\intextsep}{\edgekernelsavedintextsep}

\section{Conclusion}

Fragmented hardware and software ecosystems make it harder for LLMs to adapt and optimize kernels for resource-constrained edge devices. We introduce EdgeKernelBench, which combines a unified IR deployment framework with a hierarchical benchmark. It provides a common foundation for kernel generation, deployment, and evaluation across heterogeneous edge backends. We also introduce LEKA, which organizes optimization search across IR levels and uses device feedback to refine candidate programs. Our evaluation shows that model optimization performance varies with program complexity and target hardware. LEKA outperforms the compiler baseline on more tasks than direct generation or a budget-matched general coding agent across the evaluated backends. These findings also offer a perspective on how optimization responsibilities are divided between models and compilers. Lower-level program representations give models more direct control over hardware but introduce more complex constraints and decisions. As models become more capable, an open question is whether this boundary can shift toward lower levels. A related question is how to define layer boundaries and design IRs that help agents reason about optimization choices and constraints. EdgeKernelBench and LEKA provide an experimental foundation for studying this relationship between model capability, program abstraction, and optimization freedom.

\bibliography{references}
\bibliographystyle{iclr2027_conference}

\clearpage
\appendix
\section*{Appendix Overview}

The appendix provides supporting details in seven sections.

\begin{list}{}{\setlength{\leftmargin}{1.8em}\setlength{\labelwidth}{1.3em}\setlength{\labelsep}{0.5em}\setlength{\itemsep}{4pt}\setlength{\parsep}{0pt}\setlength{\topsep}{4pt}}
\item[\textbf{\ref{app:benchmark-details}.}] \textbf{Benchmark Specification and Evaluated Models.} Task coverage and provenance, PyTorch semantic templates and shape profiles, and the evaluated LLMs and their inference settings.

\item[\textbf{\ref{app:experimental-details}.}] \textbf{Devices and Evaluation Protocols.} Hardware and compiler configurations, correctness checks, timing boundaries, paired measurement stability, and agent evaluation tasks and budgets.

\item[\textbf{\ref{app:baseline-details}.}] \textbf{Baseline Construction and Fairness Controls.} Baseline configurations for TVM, MNN, and TensorRT, including tuning budgets, task coverage, and performance comparison settings.

\item[\textbf{\ref{app:unified-ir-framework}.}] \textbf{Unified IR and Deployment Framework.} The optimization scope of the unified IR, the generation contract and program construction, and compiler integration for backend deployment.

\item[\textbf{\ref{app:failure-cases}.}] \textbf{Failure Analysis and Case Studies.} Failure categories and their distribution across models and backends, followed by two one-shot generation cases and two LEKA optimization cases.

\item[\textbf{\ref{app:generation-prompts}.}] \textbf{Generation and Agent Prompt Templates.} The one-shot generation prompt composition, the Pure Claude Code prompt, and LEKA's shared contract and prompts for L0, L1, L2, and the hardware-sensing agent.
\item[\textbf{\ref{app:future-directions}.}] \textbf{Limitations and Future Work.} Compiler integration, coordination between IR layers and agents, and extensions to NPU and heterogeneous execution.
\end{list}

\section{Benchmark Specification and Evaluated Models}
\label{app:benchmark-details}

\subsection{Benchmark Task Coverage and Provenance}
\label{app:task-coverage-provenance}

EdgeKernelBench contains 331 canonical tasks across three complexity levels. Level~1 comprises 128 \emph{Basic Compute Patterns}, Level~2 comprises 162 \emph{Composed Operator Graphs}, and Level~3 comprises 41 \emph{Compact Model Graphs}. We further organize the tasks into mobile and personal intelligence, embodied robotics, and smart AIoT applications. This application-based grouping accounts for the diversity of workloads and deployment requirements across fragmented edge platforms. Using this application-based organization, we instantiate evaluation tasks for the scenarios represented by each platform. The resulting evaluation sets contain 91/93/20 task instances on CPU/LLVM, 98/95/25 on GPU/CUDA, and 98/105/25 on GPU/OpenCL for Levels~1, 2, and 3, respectively.

The benchmark combines manually authored tasks with implementations rewritten and adapted from selected representative open-source libraries. For tasks derived from open-source libraries, we document the original source and the changes made during adaptation. Before inclusion, each task is normalized to a static, self-contained PyTorch interface and checked for importability, interface completeness, and consistent shape materialization.

Table~\ref{tab:representative-task-coverage} summarizes representative tasks at each complexity level.

\begin{table}[!t]
\centering
\caption{\textbf{Representative tasks in EdgeKernelBench.}}
\label{tab:representative-task-coverage}
\small
\renewcommand{\arraystretch}{1.18}
\begin{tabular}{@{}>{\centering\arraybackslash}m{0.11\linewidth}
  @{\hspace{1.2em}}>{\raggedright\arraybackslash}m{\dimexpr0.89\linewidth-1.2em\relax}@{}}
\toprule
\textbf{Level} & \textbf{Representative tasks} \\
\midrule
\textbf{Level 1} &
Matrix multiplication, convolution, reductions, pooling, tensor reshaping and transposition, LayerNorm, RMSNorm, GELU, SwiGLU, rotary position embeddings (RoPE), KV cache updates, and matrix multiplication with dequantized weights. \\
\cmidrule(lr){1-2}
\textbf{Level 2} &
Convolutional blocks with normalization and activation, residual blocks, multiscale feature fusion, image and point-cloud feature fusion, RMSNorm followed by QKV projection, grouped-query attention (GQA) with KV caching, SwiGLU feed-forward blocks, and LoRA linear layers. \\
\cmidrule(lr){1-2}
\textbf{Level 3} &
MobileNetV2/V3, ResNet-18, and compact ConvNeXt classifiers; NanoDet detection networks, U-Net segmentation networks, and PointPillars backbones; Transformer decoder blocks for prefill and decoding (Llama, Qwen2, Mistral, and Phi-3); stacked TinyLlama decoder blocks. \\
\bottomrule
\end{tabular}
\end{table}

\subsection{Task Templates and Shape Profiles}
\label{app:task-templates}

Each task consists of a PyTorch semantic template and a static shape profile tailored to its application domain. The semantic template defines the module structure, reference computation, and initialization interface. The shape profile specifies input dimensions and workload parameters that reflect the computational scale and deployment needs of the target application.

\noindent\begin{minipage}{\linewidth} \textbf{Schematic PyTorch semantic template.}

\begin{lstlisting}[style=edgeprompt]
import torch
import torch.nn as nn

class Model(nn.Module):
    def __init__(self, <INITIALIZATION_ARGUMENTS>):
        super().__init__()
        <STATIC_MODEL_STATE_AND_LAYERS>

    def forward(self, <DYNAMIC_INPUTS>):
        <REFERENCE_COMPUTATION>
        return <OUTPUT_STRUCTURE>

dim_0 = {{DIMENSION_0}}
dim_1 = {{DIMENSION_1}}
configuration_0 = {{CONFIGURATION_0}}

def get_inputs():
    return [<INPUT_TENSORS_USING_STATIC_WORKLOAD_FIELDS>]

def get_init_inputs():
    return [<INITIALIZATION_ARGUMENTS>]
\end{lstlisting}
\end{minipage}

The template defines the reference computation through \texttt{Model}, supplies inputs through \texttt{get\_inputs()}, and provides model initialization arguments through \texttt{get\_init\_inputs()}.

\noindent\textbf{Schematic static shape-profile template.}

\begin{lstlisting}[style=edgeprompt]
<TASK_TEMPLATE_KEY>:
  DIMENSION_0: <STATIC_EXTENT>
  DIMENSION_1: <STATIC_EXTENT>
  CONFIGURATION_0: <STATIC_SCALAR_OR_SEQUENCE>
  scenario_note: <OPTIONAL_APPLICATION_PROVENANCE>
\end{lstlisting}
Static shape profiles are organized by application scenario so that a shared computation can be evaluated at different workload scales across edge use cases. In the profile above, \texttt{DIMENSION\_0} and \texttt{DIMENSION\_1} specify tensor dimensions, \texttt{CONFIGURATION\_0} sets a workload parameter, and \texttt{scenario\_note} records the application context.

\subsection{Evaluated LLMs and Inference Settings}
\label{app:evaluated-llms}

We evaluate DeepSeek-V4-Pro-0813, GLM-5.3, GPT-5.6-sol, Gemini-3.7-Flash, and Claude Sonnet~5 in the one-shot generation experiments. Due to inference budget constraints, both LEKA and the Pure Claude Code baseline use GLM-5.3-Flash for iterative optimization. All models use their providers' default inference settings.
\section{Devices and Evaluation Protocols}
\label{app:experimental-details}

\subsection{Systems and Compiler Configuration}

\noindent\textbf{Hardware platforms.} We evaluate three platforms representing CPU/LLVM, GPU/CUDA, and GPU/OpenCL execution. CPU experiments use a Vivo V2403A/iQOO Neo9S Pro+ with a Snapdragon~8~Gen~3 processor running Android~14. CUDA experiments run on an NVIDIA Jetson AGX Orin 64GB with Jetson Linux and CUDA~12.6. OpenCL experiments use the Mali-G610 GPU on a Firefly ROC-RK3588S-PC platform.

\noindent\textbf{Compiler integration.} We use TVM~0.23.0 with modifications to connect LLM-generated programs to compilation and deployment. EdgeKernelBench handles task import and program optimization, while TVM provides backend lowering and runtime support. The three execution paths use LLVM for CPU code generation, CUDA for the Jetson GPU, and OpenCL for the RK3588 GPU.

\subsection{Correctness and Timing Protocol}

We check each generated program against its PyTorch reference using inputs drawn from three random seeds. Correctness requires matching output structures, shapes, and data types, with numerical differences within the task-specific tolerance. Only programs that pass these checks proceed to performance evaluation.

Our paired evaluation protocol measures candidate and baseline programs in the same device session with identical inputs and matched execution settings. Timing covers one complete execution of the task, with inputs already placed on the target device. Compilation and deployment-time preparation are excluded, while computation that depends on dynamic inputs remains within the timed execution. Candidate and baseline programs use the same timing boundary within each comparison. For comparisons against TensorRT, CUDA events measure device-side latency. GPU measurements use explicit synchronization to account for execution completion.

Both programs are warmed up before repeated measurement, with warm-up settings matched within each backend. For CPU evaluation, we keep worker threads on designated cores and repeat each execution to improve timing consistency. Jetson runs in MAXN mode with its GPU clock verified at approximately 1.3\,GHz, while the RK3588 GPU is locked at 1\,GHz.

\subsection{Timing Stability and Paired Measurement}
\label{app:timing-stability}

We use the robust coefficient of variation (rCV) to assess the stability of repeated measurements. We define a unified stability criterion of rCV no greater than 5\% for candidate latency, baseline latency, and paired speedup. For a positive measurement sequence $x$, rCV is the scaled median absolute deviation divided by the sample median,
\begin{equation}
\operatorname{rCV}(x)
=\frac{1.4826\,\operatorname{median}_{i}\!\left|x_i-\operatorname{median}(x)\right|}
{\operatorname{median}(x)}.
\label{eq:timing-rcv}
\end{equation}
This measure limits the influence of isolated outliers when estimating relative timing variability.

Paired measurements balance the execution order of candidate and baseline programs to reduce bias from device-state drift. For each retained pair $i$, let $t_{\mathrm{baseline},i}$ and $t_{\mathrm{candidate},i}$ denote the baseline and candidate latencies. We compute $s_i=t_{\mathrm{baseline},i}/t_{\mathrm{candidate},i}$ and report the median paired speedup. We assess paired stability by computing rCV over $\{s_i\}$ alongside the two latency sequences. This checks the repeatability of both individual timings and the relative performance comparison.

\subsection{Agent Evaluation Tasks and Budgets}
\label{app:agent-evaluation-settings}

\noindent\textbf{Task selection.} Iterative agent optimization requires substantially more evaluation time per task than direct generation. For the agent comparison in Section~\ref{sec:agent-evaluation}, we select a subset by stratified random sampling across complexity levels. We sample 23 tasks per backend, comprising 10 Level~1, 9 Level~2, and 4 Level~3 tasks, for 69 task instances in total. This allocation retains all three optimization scopes within the available evaluation budget.

\noindent\textbf{Optimization budgets.} LEKA and Pure Claude Code receive identical optimization time budgets for each task. The per-task upper limits are 90 minutes for Level~1 and 155 minutes for Levels~2 and~3. These limits bound the time available for iterative generation and evaluation on each task.

\noindent\textbf{Performance aggregation.} Following the metrics used in the main experiments, we report F@1 relative to TVM and GM relative to Pure Claude Code for the final valid programs. F@1 measures the fraction of evaluated tasks with correct implementations and valid timings that outperform the TVM baseline. For Figure~\ref{fig:leka-results}(c), GM aggregates the per-task ratios of Pure Claude Code latency to LEKA latency, including tasks where LEKA is slower. A GM above one indicates an overall speed advantage for LEKA on the evaluated task set.

\section{Baseline Construction and Fairness Controls}
\label{app:baseline-details}

\subsection{TVM Compiler Baseline}
\label{app:tvm-baseline}

\noindent\textbf{Budgeted compiler reference.} We construct the TVM baseline by importing each PyTorch reference and extracting the optimization tasks for MetaSchedule. Before evaluation, we search and measure candidate schedules on the target hardware using the convergence-calibrated budgets described below. The resulting tuning database stores the schedules and their measured latencies for each workload and target. We build the baseline from the lowest-latency valid schedule found for the matching workload and target within this search budget.

\noindent\textbf{Backend-specific calibration.} A benchmark task may yield multiple MetaSchedule tuning tasks, each corresponding to a PrimFunc exposed after graph lowering and fusion. Since convolution, matrix multiplication, reduction, and elementwise programs induce substantially different search spaces, a single fixed trial count would allocate the same cost to tasks with very different scheduling difficulty. We group the extracted tasks into the six classes in Table~\ref{tab:tvm-ms-task-classes} according to their dominant computation patterns and associated search complexity. For each backend and class, we select representative PrimFuncs with different input shapes, covering smaller and larger computation sizes where available. Each configuration is tuned independently with two random seeds. During calibration, the isolated PrimFunc receives the entire preset ceiling: 1,000 hardware trials on CPU/LLVM and 1,500 on the two GPU backends. All runs use an XGBoost cost model, the gradient task scheduler, and \texttt{num\_trials\_per\_iter}=64.

\begin{table}[!t]
\centering
\caption{\textbf{Search-complexity classes used for MetaSchedule budget calibration.} The classes summarize the dominant computation patterns and scheduling structures of the extracted tasks.}
\label{tab:tvm-ms-task-classes}
\footnotesize
\setlength{\tabcolsep}{3.8pt}
\renewcommand{\arraystretch}{1.08}
\begin{tabular*}{\linewidth}{@{\extracolsep{\fill}}c
  >{\raggedright\arraybackslash}p{0.39\linewidth}
  >{\raggedright\arraybackslash}p{0.43\linewidth}@{}}
\toprule
\textbf{Class} & \textbf{Matching rule / family} & \textbf{Dominant search structure} \\
\midrule
C1 & \texttt{conv} & Convolutional tiling, layout, and data reuse \\
C2 & \texttt{matmul|dense|linear|bmm} & Matrix-multiplication tiling and reduction mapping \\
C3 & \texttt{softmax|norm|mean|sum|}\newline
     \texttt{variance|pool} & Reduction, normalization, and pooling schedules \\
C4 & \texttt{lstm|gru} & Recurrent computation with loop-carried state \\
C5 & \texttt{cumsum|sort|topk} & Scan, sorting, and synchronization structure \\
C6 & Catch-all & Elementwise, activation, and data-movement programs \\
\bottomrule
\end{tabular*}
\end{table}

\noindent\textbf{Convergence-derived budgets.} Let $\ell_{b,r,s}(t)$ denote the valid latency measured at hardware trial $t$ for representative PrimFunc and shape $r$, random seed $s$, and backend $b$. The best valid latency observed through trial $t$ is
\begin{equation}
  \ell^{\star}_{b,r,s}(t)
  = \min_{1 \leq j \leq t} \ell_{b,r,s}(j),
  \label{eq:tvm-ms-incumbent}
\end{equation}
where $T_{\max}$ is the calibration trial limit and $\ell^{\star}_{b,r,s}(T_{\max})$ is the final incumbent. We define the convergence trial as
\begin{equation}
\begin{split}
  T^{\mathrm{conv}}_{b,r,s}
  = \min \bigl\{t\ \bigm|\ &\ell^{\star}_{b,r,s}(t)
      \leq (1+\epsilon)\ell^{\star}_{b,r,s}(T_{\max}),\\[-1mm]
    &\min_{t \leq j < t+W}\ell^{\star}_{b,r,s}(j)
      \geq (1-\epsilon)\ell^{\star}_{b,r,s}(t),\quad
      t+W\leq T_{\max}+1\bigr\},
  \label{eq:tvm-ms-convergence}
\end{split}
\end{equation}
with $\epsilon=0.02$ and $W=100$. Thus, the selected point is already within 2\% of the final incumbent and is followed by 100 trials without a further improvement greater than 2\%.

\begin{figure}[!h]
  \centering
  \includegraphics[width=\linewidth]{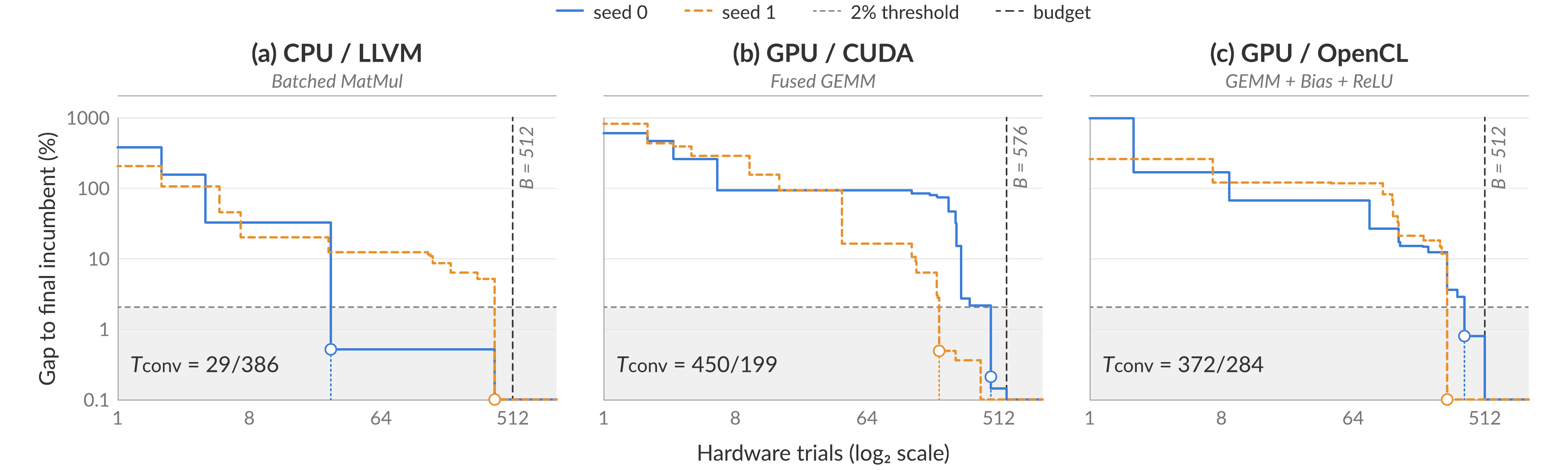}
  \caption{\textbf{MetaSchedule calibration curves.}}
  \label{fig:tvm-ms-convergence}
\end{figure}

Figure~\ref{fig:tvm-ms-convergence} illustrates calibration on three representative workloads, each evaluated with two seeds. Panels (a), (b), and (c) show batched matrix multiplication on CPU/LLVM, GEMM with bias and elementwise multiplication on GPU/CUDA, and GEMM with bias and ReLU on GPU/OpenCL, respectively. Their input shapes are $(8,48,48)\times(8,48,129)$, $(6,592)\times(592,296)$, and $(4,128)\times(128,8)$. Each curve tracks the percentage gap between the best-so-far latency and the final best latency for that seed, with both axes on logarithmic scales. Open markers indicate convergence, the shaded region marks gaps below 2\%, and vertical dashed lines indicate the adopted budgets. Zero gaps are displayed at 0.1\% to accommodate the logarithmic axis.

For class $c$, let $\mathcal{R}_{b,c}$ contain all successful representative PrimFunc--shape--seed configurations used on backend $b$. Its nominal per-PrimFunc budget is
\begin{equation}
  B_{b,c}
  = \max\!\left\{
      64,
      64\left\lceil
        \frac{1.25\max_{(r,s)\in\mathcal{R}_{b,c}}
        T^{\mathrm{conv}}_{b,r,s}}{64}
      \right\rceil
    \right\}.
  \label{eq:tvm-ms-class-budget}
\end{equation}
The 25\% reserve absorbs variation across shapes and search seeds, while rounding to a multiple of 64 aligns the cap with the hardware-measurement batch. Figure~\ref{fig:tvm-ms-convergence} shows the C2 representatives that determine the reported budget on each evaluated backend.

\noindent\textbf{Benchmark task allocation.} For each benchmark task, we set \texttt{max\_trials\_global} to the sum of the class budgets assigned to all extracted MetaSchedule tasks and set \texttt{max\_trials\_per\_task} to the largest of those budgets, while retaining \texttt{num\_trials\_per\_iter}=64. The gradient task scheduler allocates this total budget adaptively among the extracted tasks, subject to the shared per-task limit. The total search budget therefore reflects both the number and the computational characteristics of the PrimFuncs extracted from a benchmark task.

\noindent\textbf{Optimization cost.} MetaSchedule search repeatedly compiles and measures candidate schedules on the target device. This repeated compilation and hardware measurement adds optimization time as the number of tasks and candidate schedules grows. Direct LLM generation avoids this repeated online schedule search. The generated program is then compiled, validated, and measured. Table~\ref{tab:tvm-baseline-results} shows that one-shot generation can outperform the tuned TVM baseline on a subset of tasks. These results suggest a route to reducing hardware search costs while producing efficient implementations.

\noindent\textbf{Fairness controls.} TVM and LLM-generated programs follow the correctness and paired measurement protocol in Appendix~\ref{app:experimental-details}. We compare freshly measured execution times without applying additional MetaSchedule tuning to the LLM-generated programs.

\FloatBarrier

\subsection{MNN Inference-Engine Baseline}

\noindent\textbf{Fixed inference-engine reference.} To compare generated programs with implementations from a mature edge inference engine, we construct a baseline using MNN~3.3.0. Each PyTorch reference program is first exported to ONNX opset~17 and then converted to an MNN model under a common unquantized FP32 numerical contract. We enable MNN's speed-preferred graph optimization and execute the resulting model through the native target backend. Execution uses the hand-optimized operator implementations maintained within MNN, together with its graph optimizer and backend scheduler. For multi-operator Level~2 and Level~3 tasks, we measure one complete execution of the workload graph; operator partitioning and fusion performed internally by MNN are both treated as valid inference-engine implementations.

\noindent\textbf{Evaluation coverage.} We evaluate MNN on the backends and workloads it supports, with correctness verified against the PyTorch references. The comparison covers 176 CPU/LLVM tasks and 203 GPU/OpenCL tasks. The Level~1, 2, and 3 counts are 73/84/19 for CPU/LLVM and 79/101/23 for GPU/OpenCL. Our MNN evaluation covers CPU and OpenCL execution, while TensorRT provides the inference-engine comparison on CUDA. All LLMs use the same task set within each backend to compute CSR, FCR, and F@1.

\noindent\textbf{Paired performance measurement.} We compare MNN with LLM-generated programs using the correctness and paired timing protocol in Appendix~\ref{app:experimental-details}.

\subsection{TensorRT Inference-Engine Baseline}
\label{app:tensorrt-baseline}

\noindent\textbf{Baseline construction and precision settings.} We use TensorRT~10.3.0 to construct the GPU/CUDA inference-engine baseline on the Jetson AGX Orin 64GB. Each PyTorch reference program is exported through ONNX opset~17, after which TensorRT optimizes the graph and builds the engine. TensorRT retains the data types and type conversions defined in each reference program. Quantization, TF32, and automatic precision reduction are disabled. TensorRT performs constant folding, operator fusion, and algorithm selection at the original precision. We use TensorRT's built-in implementations without adding external custom plugins or handwritten kernels. For Level~2 and Level~3 tasks, we deploy and measure the complete computation graph, allowing TensorRT to choose its internal fusion and execution plan.

\noindent\textbf{Task coverage and correctness validation.} We evaluate TensorRT on supported tasks that pass the correctness checks described in Appendix~\ref{app:experimental-details}. This yields 215 of the 218 GPU/CUDA tasks. Levels~1, 2, and 3 contain 95, 95, and 25 supported tasks, respectively.

\noindent\textbf{Paired performance measurement.} We compare TensorRT with LLM-generated programs using the correctness and paired timing protocol in Appendix~\ref{app:experimental-details}.

\section{Unified IR and Deployment Framework}
\label{app:unified-ir-framework}
Section~\ref{sec:edge-pipeline} introduces the shared representation and evaluation pipeline. Here we describe the optimization scope, generation contract, and compiler integration.

\subsection{Unified IR and Optimization Scope}
Unified Kernel IR describes deployment programs at two complementary levels. The graph and invocation level expresses computational dependencies, fusion boundaries, and the flow of intermediate results. The computation and scheduling level specifies loops, data layouts, memory accesses, and parallel mappings. Together, these levels let the model coordinate local kernel changes with the organization of the surrounding program.

Memory decisions span both levels. Fusion and decomposition determine which intermediate tensors are materialized and how long they remain live. Local scheduling determines data reuse and temporary storage within each computation. Exposing these choices allows the model to weigh execution latency against memory use under the target device's resource constraints. The representation accommodates different parallel strategies and storage organizations for different targets.

\subsection{Generation Contract and Program Construction}
The PyTorch reference defines the computation, while input and output specifications and numerical tolerances define the program's observable behavior. Within this contract, the LLM can reorganize computations, intermediate layouts, fusion boundaries, and execution schedules. These choices preserve the task interface while allowing different implementations of the same computation.

The framework builds an executable program from the generated artifact and checks its structure and compatibility with the target backend. Numerical validation then verifies that its outputs agree with the PyTorch reference within the specified tolerances.

\subsection{Compiler Integration and Backend Deployment}
Progressive lowering converts the representation into an executable artifact through compiler transformations. These transformations resolve storage and execution details under the backend's constraints. The deployment path loads the artifact and binds its entry point to the runtime interface.

Our implementation uses TVM's layered IR, code generation, and runtime support for program construction and device execution. Within this pipeline, EdgeKernelBench provides task and generation interfaces through which LLMs optimize operator implementations, then passes the programs to the evaluator. Backend integration lowers LLM-generated programs from the unified IR to the target backend without introducing additional optimization. The lowered programs follow compilation and runtime paths for CPU/LLVM, GPU/CUDA, or GPU/OpenCL. These components provide backend integration and runtime support for the generated programs.

\subsection{Compiler Instantiation and Hardware Adaptation}

When instantiating EdgeKernelBench with a new compiler, we implement the unified generation interface using that compiler's IR. The LLM then writes operator implementations and expresses optimization decisions in this representation. Compiler adapters connect the generated programs to the corresponding lowering, code generation, and runtime paths. For new hardware supported by the compiler, we extend the target configuration and deployment adapter to connect to the device runtime. These extensions retain the same task definitions and evaluation criteria.

\section{Failure Analysis and Case Studies}
\label{app:failure-cases}

\subsection{Backend- and Model-Stratified Failure Analysis}
\label{app:stratified-failure-analysis}

We extend Section~\ref{sec:error-analysis}'s analysis with error definitions and comparisons across models and backends. Figure~\ref{fig:backend-failure-composition} shows the failure composition of one-shot generation for each model and backend.

\textbf{Syntax/API} includes incomplete source responses, invalid program syntax, undefined symbols, unavailable interfaces, and malformed program structure. \textbf{Memory} covers illegal buffer-allocation scopes, pointer or address-space declarations, and invalid memory accesses; \textbf{Parallel} includes missing thread bindings, invalid launch or work-group mappings, and synchronization patterns rejected by backend legality checks. \textbf{Shape/type} captures violations of tensor rank, broadcasting, reshape, data-type, or vector-width contracts, whereas \textbf{Correctness} denotes output-value or consistency mismatches. \textbf{Other} covers timeouts and remaining compiler, execution, or verification exceptions.

\noindent\begin{minipage}{\linewidth}
    \centering
    \includegraphics[width=\linewidth]{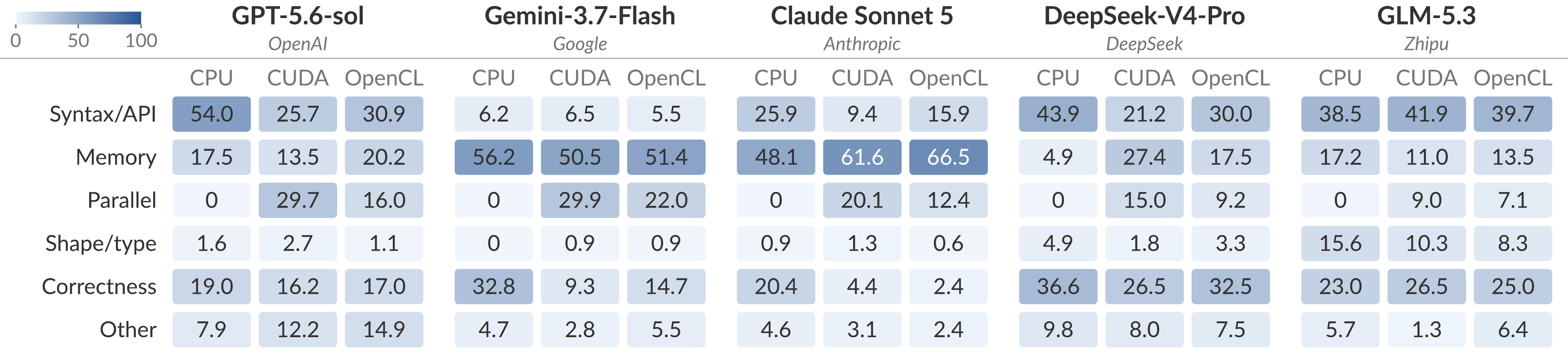}
    \makeatletter\def\@captype{figure}\makeatother \caption{\textbf{Failure composition by model and backend.} Each model group contains CPU/LLVM, GPU/CUDA, and GPU/OpenCL subcolumns; rows indicate error categories. Each subcolumn reports the composition of unsuccessful one-shot outcomes and sums to 100.0\% before rounding. Zero denotes no observed failures in that category, including CPU/LLVM Parallel entries. All groups share the upper-left color scale; percentages describe failure composition rather than failure rates.}
    \label{fig:backend-failure-composition}
\end{minipage}
\par

On CPU/LLVM, Syntax/API accounts for 54.0\% of GPT-5.6-sol failures, while Memory accounts for 56.2\% of Gemini-3.7-Flash failures. This contrast points to different obstacles in constructing valid programs within the same execution environment. The former concerns expressing computations through valid syntax and interfaces, while the latter concerns buffer scope and legal memory access. Memory remains the largest failure category for Gemini-3.7-Flash and Claude Sonnet 5 on both GPU backends, accounting for more than half of their failures in each case. The persistence of this pattern across backends suggests that memory management is a recurring difficulty for these models. These results indicate persistent limitations in LLMs' memory management capabilities.

For GPT-5.6-sol, Parallel is the largest failure category on CUDA at 29.7\%, while Syntax/API leads on OpenCL at 30.9\%. The CUDA failures point to difficulties in satisfying thread-mapping and synchronization constraints, whereas the OpenCL results place greater emphasis on valid program construction. These contrasting profiles favor backend-specific feedback that identifies which constraints a candidate violates. Numerical correctness also remains a challenge, accounting for 32.5\% of DeepSeek-V4-Pro failures on OpenCL. These programs compile and execute but fail to reproduce the reference outputs within the required tolerances. This gap points to limitations in preserving computational semantics when translating reference computations into optimized implementations. Overall, these results reveal limitations in LLMs' ability to construct valid programs, manage memory and parallel execution, and preserve computational semantics.

\subsection{Optimization Case Studies}
\label{app:optimization-cases}

\subsubsection{One-shot Generation Case Studies}
\label{app:oneshot-cases}

We examine two one-shot implementations through generated code and paired measurements following the protocol in Appendix~\ref{app:experimental-details}.
\paragraph{Selective fusion and intermediate storage.}
On the OpenCL ConvNeXt block, Gemini-3.7-Flash reduces latency from 18.0\,ms to 0.6\,ms, achieving a $30.9\times$ speedup over TVM. The implementation uses separate kernels for depthwise convolution and channel transformations to accommodate their different memory access patterns. The convolution kernel reuses spatial neighborhoods in workgroup memory and writes its output in the layout required by the next stage. The second kernel fuses normalization, two linear transformations, activation, and the residual connection. It avoids writing the full expanded hidden activation to global memory. For this task, GPT-5.6-sol generates one kernel that fuses the entire block. Its implementation tiles the hidden channels and repeatedly synchronizes the workgroup between tile computations. It achieves only $0.9\times$ the performance of its corresponding TVM baseline. This comparison suggests that extensive fusion does not necessarily yield better performance. Separating stages with different access patterns can improve data reuse and reduce synchronization overhead, even when it introduces an intermediate tensor.

\paragraph{Joint optimization of dequantization and matrix multiplication.}
On the CPU dequantization and matrix multiplication task, Gemini-3.7-Flash reduces latency from 39.7\,ms with MNN to 2.1\,ms. The paired speedup is $19.0\times$. The reference program converts integer weights to floating point, applies zero points and scales, and then performs matrix multiplication. The generated implementation reuses the reference program's integer weights and per-channel quantization parameters. It moves zero-point correction and scaling after the reduction, reading integer weights directly for floating-point accumulation. This rearrangement avoids creating and storing the entire weight matrix in floating-point form. The implementation parallelizes blocks of output channels and uses vectorized loops to access contiguous weights and accumulate outputs. For the $2048\times5632$ weight matrix, it avoids a 44\,MiB floating-point intermediate. The case illustrates how algebraic rearrangement can reduce both intermediate storage and elementwise work within the reduction.

\subsubsection{LEKA Optimization Case Studies}
\label{app:leka-cases}

We trace two LEKA runs with GLM-5.3-Flash to examine how optimization proceeds across program layers. The final implementations are compared against the tuned TVM baselines using the evaluation protocol in Appendix~\ref{app:experimental-details}.

\paragraph{From graph rewrites to device-level refinement.}
On the CUDA task combining batched matrix multiplication, instance normalization, addition, and multiplication, LEKA achieves a latency of 16.2\,$\mu$s in formal evaluation. The final implementation achieves a $1.2\times$ speedup over the tuned TVM baseline at 19.8\,$\mu$s. This computation mixes matrix arithmetic and reductions with short elementwise operations, leaving optimization opportunities at several program layers.

L0 first fuses the final elementwise operations. It uses the parameter preparation mechanism to move fixed-weight layout conversion and normalization constant preparation outside the repeated inference call. Given the weight layout, L1 fuses matrix multiplication with bias addition and reschedules the matrix computation and normalization reduction. These changes reduce the search-stage latency of this branch from approximately 104.0\,$\mu$s to 20.0\,$\mu$s. L2 redistributes work within thread blocks to avoid duplicate weight loads and uses vectorized accesses and register reuse to reduce memory traffic. Warp-level reductions also reduce block-wide synchronization in the normalization kernel from eight barriers to one. The resulting implementation runs in approximately 16.2\,$\mu$s.

The trajectory reveals dependencies between decisions. The graph layer establishes the parameter layout and computation boundaries that the scheduling layer uses to organize parallel work. Source-level refinement improves memory access and synchronization within it. After an upstream change, LEKA rebuilds the affected downstream implementations and retains the best verified complete program. This preserves compatible improvements as search moves between layers.

\paragraph{Hardware feedback directs schedule refinement.}
On the CPU task combining convolution with addition, scaling, sigmoid, and group normalization, LEKA reaches 17.0\,ms in formal evaluation. This is a $1.6\times$ speedup over the TVM baseline at 27.5\,ms. Although the graph contains several elementwise and reduction stages, convolution still accounts for approximately 85.0\% of the profiled interval after the first effective scheduling pass. This evidence directs further optimization toward data reuse and register blocking in the convolution.

L1 constructs a convolution microkernel that computes several output channels and adjacent spatial positions together. It retains accumulations in registers and uses loop unrolling and vectorization to reuse inputs. This optimization reduces the search-stage latency to 19.7\,ms. With profiling showing that convolution remains dominant, L1 expands the microtile from eight output channels by four spatial positions to eight channels by eight positions. The wider spatial tile amortizes each weight load over more multiply-accumulate operations while keeping the accumulators within the register budget. This refinement lowers the search-stage latency to approximately 17.0\,ms.

Hardware feedback identifies the dominant computation, and L1 refines its loops while preserving the graph and call interfaces. The implementation preserves the L0 graph and achieves its gains through L1 scheduling without further L2 code changes. The case illustrates how the layered design supports sustained optimization within the layer that contains the measured bottleneck.

\raggedbottom
\section{Generation and Agent Prompt Templates}
\label{app:generation-prompts}

This section describes the prompts used for direct generation and agent-based optimization. Task subsets, scoring conventions, and agent budgets are specified in Appendix~\ref{app:agent-evaluation-settings}.

\subsection{One-Shot Generation Prompt Protocol}
\label{app:single-query-prompt}

We outline the prompt structure and the instructions shared across tasks. The prompt introduces the optimization objective and program requirements, provides device context and one demonstration, and concludes with the target task. We group device capabilities and backend constraints under a single \emph{Device Hardware Profile} in the template below. The demonstration illustrates the expected implementation format, while the target task supplies the PyTorch reference and tensor specifications. One-shot generation returns one candidate without iterative evaluation feedback.

\begin{lstlisting}[style=edgeprompt]
SYSTEM MESSAGE
<SYSTEM_INSTRUCTIONS>
optimization objective; program representation and syntax guidance;
correctness requirements; response format and evaluation interface
</SYSTEM_INSTRUCTIONS>
<DEVICE_HARDWARE_PROFILE>
device capabilities; resource constraints; backend execution support
</DEVICE_HARDWARE_PROFILE>

DEMONSTRATION USER MESSAGE
<DEMONSTRATION_TASK>
one-shot example task and reference program
</DEMONSTRATION_TASK>

DEMONSTRATION ASSISTANT MESSAGE
<DEMONSTRATION_RESPONSE>
one complete and validated implementation
</DEMONSTRATION_RESPONSE>

TARGET USER MESSAGE
<TASK_SPECIFICATION>
initialization arguments; input, state, and output specifications
</TASK_SPECIFICATION>
<PYTORCH_REFERENCE>
PyTorch reference program
</PYTORCH_REFERENCE>
\end{lstlisting}

The following template summarizes the core generation instructions.

\begin{lstlisting}[style=edgeprompt]
TASK
Given a fixed-shape PyTorch reference and its tensor specifications,
produce one complete, executable, and efficient program. Preserve
the reference computation and the evaluation interface.

You may reorganize computation, change operator boundaries, fuse
operations, alter data layouts, choose equivalent algorithms, and
apply other implementation strategies that improve execution
efficiency while preserving the task semantics.

TASK INFORMATION
Use the reference program and public input, state, and output
specifications to guide implementation. Static parameters may be
prepared during deployment. Computation that depends on dynamic
inputs belongs in the measured program.

CORRECTNESS
Match the reference output specifications and numerical tolerances.
Only correct implementations are evaluated for performance.

OUTPUT
Return one complete implementation compatible with the evaluation
interface, without explanatory text or incomplete code.

<TASK_SPECIFICATION>
{initialization arguments; input, state, and output specifications}
</TASK_SPECIFICATION>

<PYTORCH_REFERENCE>
{PyTorch reference program}
</PYTORCH_REFERENCE>
\end{lstlisting}

Generation and evaluation run in isolated environments to prevent access to hidden evaluation data or interference with scoring.

\subsection{Agent Prompt Protocols}
\label{app:prompt-protocols}

Agent optimization uses the same public reference and tensor specifications as direct generation, with iterative evaluation within the assigned budget. The templates below describe the optimization roles and their interaction with the evaluator.

\subsubsection{Pure Claude Code Baseline Prompt}
\label{app:pure-claude-code-prompt}

The pure Claude Code baseline uses the same base model, public task specification, evaluator, and wall-clock budget as LEKA. A persistent Claude Code session uses the task definition and its initial high-level IR to construct and refine complete candidates. The implementation-independent core prompt is summarized below.

\begin{lstlisting}[style=edgeprompt]
ROLE
You are the general-purpose program-optimization agent for this task.
Within the allotted time, construct, validate, and continuously
improve one complete candidate program.

TASK
Using the reference program and tensor specifications, produce an efficient
implementation that preserves the input/output interface and numerical
semantics. You may change program organization, operator boundaries,
algorithms, data layouts, and execution schedules, but every submitted
candidate must be complete and executable.

AVAILABLE CONTEXT
<TASK_DEFINITION>
{PyTorch reference and public input, state, and output specifications}
</TASK_DEFINITION>
<INITIAL_IR>
{high-level IR derived from the task definition}
</INITIAL_IR>

OPTIMIZATION PROCESS
Construct the first complete candidate from the public task materials.
You may then modify candidates and invoke the controlled evaluator.
Use the resulting execution evidence to choose subsequent changes,
and associate each change with a concrete optimization hypothesis.

EVALUATION ISOLATION
Optimize within the provided workspace using public task information
and authorized feedback. Keep hidden evaluation data and the scoring
process isolated from candidate construction and execution.

COMPLETION
Continue until further improvement is unlikely or the wall-clock
budget expires. Leave one complete, validated best candidate for final
evaluation by the trusted evaluator.
\end{lstlisting}

\subsubsection{LEKA Hierarchical Agent Prompts}
\label{app:leka-prompts}

LEKA decomposes optimization along the IR hierarchy and assigns each editable representation to a resident agent. Each agent explores the optimization space exposed by its representation and coordinates changes with adjacent layers through shared evaluation feedback. The hardware-sensing agent remains read-only and converts measurement evidence into layer-specific optimization directions. The following templates summarize the shared instructions and each role's optimization scope.

\noindent\textbf{Shared layer instructions.}

\begin{lstlisting}[style=edgeprompt]
ROLE
You are a resident optimizer for one layer of a hierarchical program.
Explore the optimization space exposed by your layer's IR, using
adjacent representations as context for coordinated improvements.

INPUT
You receive the task definition, current layer program, relevant
context from adjacent layers, evaluation evidence, and cross-layer
optimization advice.

PROTOCOL
Preserve task semantics and compatibility between program layers.
Use evaluation evidence to identify an optimization opportunity and
explain how the proposed change addresses it. Validate correctness
and measure performance through the common evaluator.

Express changes within your layer's optimization scope. Coordinate
changes beyond that scope through a request to the relevant agent,
with supporting evidence and the intended optimization.

OUTPUT
Return one complete candidate for your layer, or NOOP if no justified
change is available. Candidate adoption requires verified correctness
and an improvement in the measured objective.
\end{lstlisting}

\noindent\textbf{L0: graph and algorithm agent.}

\begin{lstlisting}[style=edgeprompt]
TASK
Optimize the complete high-level program using the public reference,
task specifications, and current program representation.

SCOPE
You may change the algorithmic expression, graph topology, global
fusion or decomposition boundaries, data layout, and deterministic
deployment-time preparation of static model state.

BOUNDARY
Express optimizations through the graph and algorithm choices
available in the high-level IR. Preserve the task interface and
numerical semantics so that subsequent layers can refine the program.

OUTPUT
Return one complete high-level program that can be lowered and
evaluated through the common interface.
\end{lstlisting}

\noindent\textbf{L1: loop, scheduling, and local-composition agent.}

\begin{lstlisting}[style=edgeprompt]
TASK
Optimize the normalized compute representation derived from the
current high-level program.

SCOPE
You may transform loop structure, parallel mapping, memory hierarchy,
data access, and scheduling. You may also jointly optimize local
operator regions supported by the scheduling representation.

BOUNDARY
Express changes within the scheduling IR's optimization space while
preserving external interfaces and the surrounding graph. Coordinate
broader graph changes with L0 and target-source refinements with L2.

OUTPUT
Return one complete scheduling-layer program, together with any
cross-layer request required by the proposed optimization.
\end{lstlisting}

\noindent\textbf{L2: target-source agent.} The CUDA, OpenCL, and LLVM instances share the following instructions, supplemented by context for their target representation and execution constraints.

\begin{lstlisting}[style=edgeprompt]
TASK
Optimize the target-source implementation deterministically produced
from the current upper-layer program.

SCOPE
Within the existing call interface, you may rewrite compute bodies and
optimize instruction organization, vectorization, local data access,
and other target-specific implementation details.

BOUNDARY
Explore the implementation choices exposed by the target-source
representation while preserving compatibility with upper layers and
the runtime interface. Coordinate changes to graph structure,
loop organization, or launch configuration with L0 or L1.

OUTPUT
Return one complete target-source candidate suitable for interface
validation, hidden correctness testing, and latency measurement.
\end{lstlisting}

\noindent\textbf{Hardware-Sensing Agent.}

\begin{lstlisting}[style=edgeprompt]
ROLE
You are a read-only hardware-evidence analyst who guides optimization
through measured program behavior.

INPUT
You receive trusted measurement summaries, the current program stack,
and the relevant optimization history.

TASK
Identify a performance bottleneck supported by the available evidence,
determine whether it is most directly actionable at L0, L1, or L2,
and propose a testable optimization direction for that layer.
Distinguish measured observations from hypotheses to be evaluated.

OUTPUT
Return the target layer, supporting observations, confidence, and
suggested optimization. State when the evidence is insufficient to
recommend a change.
\end{lstlisting}

These prompts organize optimization around each representation's available transformations and use evaluation feedback to coordinate changes across layers. All candidates remain subject to the same correctness and performance evaluation.

\section{Limitations and Future Work}
\label{app:future-directions}

\subsection{Limitations}

Our evaluation covers CPU/LLVM, GPU/CUDA, and GPU/OpenCL, but does not yet include specialized accelerators such as NPUs. We evaluate each task on a CPU or a GPU separately, without partitioning its computation between the two for cooperative execution. Such settings involve task allocation and resource sharing across compute units and would broaden the evaluation scope.

\subsection{Future Work}

Future work could explore how the organization of the unified IR supports focused agent reasoning and coordinated optimization. Using multilevel compiler infrastructure such as the MLIR-based IREE,\footnote{\url{https://iree.dev/}} EdgeKernelBench could expose finer-grained representations for optimization and support different divisions of work among agents. Comparing layer boundaries, context sharing, and cross-layer feedback would help identify how to narrow each agent's search space while retaining opportunities to jointly optimize computation, memory use, and scheduling.

Extending the evaluation to NPUs would enable research on program optimization under specialized compute and on-chip memory constraints. For CPUs and GPUs within the same device, we could jointly optimize computation partitioning, data transfers, and synchronization for a single task. This would allow us to examine how the benefits of heterogeneous execution vary with workloads and resource budgets, and to explore trade-offs between performance and memory use.

\end{document}